\documentclass[letterpaper, 10 pt, conference]{ieeeconf}  

\IEEEoverridecommandlockouts                              
\usepackage{graphics} 
\usepackage{epsfig} 
\usepackage{mathptmx} 
\usepackage{amsmath} 
\usepackage{amssymb}  
\usepackage{algorithm2e}
\usepackage{color}
\usepackage[noend]{algpseudocode}
\usepackage{hyperref}
\usepackage{dblfloatfix} 

\title{\LARGE \bf
Motion Rectification for an Homeostasis-Enabling Wheel
}

\author{Juan-Pablo Afman, Mark Mote, and Eric Feron$^{1}$
\thanks{*This work was supported by the Georgia Institute of Technology and the Georgia Tech Research Institute}
\thanks{$^{1}$Pablo Afman, Mark Mote, and  Eric Feron are with the Decision and Control Laboratory, Georgia Institute of 
Technology, Atlanta, GA, USA. {\tt\small \{jafman3, mmote3, feron\} at gatech.edu}. Feron also holds 
courtesy appointments with the Institut Sup\'erieur de l'A\'eronautique et de l'Espace and the Ecole Nationale de l'Aviation Civile, Toulouse, France.}%
}

\begin{document}

\maketitle
\thispagestyle{empty}
\pagestyle{empty}

\begin{abstract}

A wheel that is capable of producing thrust and maintaining vehicle internal integrity is presented. The wheel can be seen as an organic extension to the central unit (eg the vehicle) it is attached to, that is, the system and the wheel can be completely surrounded by the same tegument while enabling continuous wheel rotation without tearing the tegument. Furthermore, a skeleton linking the central unit of the system to the wheel's center can be made through the use of joints and linear links, while allowing the apparatus to rotate continuously in the same direction with bounded twisting and no tegument tear. For that reason, artificial muscles can also be used to actuate the entire system. 
The underlying enabling mechanism is the rectification of a small number of oscillatory inputs. Another contribution of the proposed setup is to offer a plausible, yet untested, evolutionary path from today's living animals towards animals capable of wheeled locomotion.
\end{abstract}

\section{Introduction}
Homeostasis is the tendency for several key and interpendent elements of a living system to evolve toward a relatively stable equilibrium, especially as maintained by physiological processes~\cite{Ber:74,Rig:70,Can:26}. 
These properties include body temperature, pH, osmotic pressure, carbohydrate concentration in the blood, and so on. All mammals are characterized by a strong homeostasis, whereby the inner body is strongly regulated and isolated from outside conditions via the tegument (skin, nails and hair) and the mucous membranes (along lips, mouth, stomach, colon, reproductive system etc.)

The meaning of homeostasis can be extended to machines to describe the ability of some engineered systems to exercise strong inner regulations (including temperature, pressure, relative humidity) and whose internal elements are shielded away from the outside world via a comprehensive set of two-dimensional membranes that create a partition of the system's environment into two non-intersecting components: The {\em inside} of the system, which contains all the system's vital elements, the tegument and the mucous tissue and whose interior is connected~\cite{Kur:66,ROD:91}; and the {\em outside} of the system, which, informally, is the "rest of the world". Not all engineered systems feature this type of property. For example, "inside" and "outside", as defined in this paper, do not exist as separate elements in vehicles such as today's cars, ships (including submarines), turbine-run power plants etc. Some systems could be easily modified to have clearly identified "inside" and "oustide"~\cite{cubliIROS12}, but they do not have wheels (unless the entire body is shaped as a wheel).

The introduction of the wheel and other axle-mounted rotary devices has not been able to retain the notion of an "inside" whose interior is connected in engineered systems. For example, conventional wheels or propellers do not share the same "inside" as the vehicle they propel. The inability for the "central unit" of a vehicle to access some of its key components "from the inside" comes at the cost of significantly increased complexity when system maintenance or repair is necessary, and often means the system must be stopped when these operations have to be performed. By contrast, a wounded hand can be fixed "from the inside" during otherwise normal human activities. The inability to enable homeostasis is often cited as the leading reason for why nature "was not able to invent the wheel" in higher order life forms, and it is the object of several publications with various degrees of formalism~\cite{Wik:17a,Ada:98,ScD:10,Dia:83}.  Indeed, it is impossible to draw conventional wiring and actuation (nerves, blood and lymphatic vessels, muscles) from the center of the body to conventional wheels, and a path from today's higher life forms to conventionally wheeled animals would have to be drastically different from all known biological evolution mechanisms.

To "excuse" nature's inability to come up with the concept of wheel, environmental factors are often cited: For example, nature did not invent the wheel because the wheel is not adapted to environments where man-made infrastructures, such as roads, are missing. As experimental evidence, it is reported that wheeled transportation was voluntarily abandoned in North Africa thousands of years ago because of lack of adaptation of the wheel to desert transport~\cite{Dia:83}. This argument, however, does not explain why the Jet Propulsion Laboratory and the National Aeronautics and Space Administration have consistently sent wheeled rovers and robots to successfully explore the deserts of the Moon and Mars~\cite{You:07,Squ:05,Sci:12}. Concurrently, more "inclusive" publications support that the wheel, as we know it, is the result of a natural evolution process, one "cog" of which is the human species and its engineering genius~\cite{Gam:09,Bej:10,ScD:10}. 

The discussion below shows that it is possible to construct a functional wheel capable of performing the same transportation functions as conventional wheels, while respecting the topological constraints required by homeostasis in high-order living species: The system consisting of the set of wheels and its payload features a continuous, possibly elastic, membrane that completely separates the system's connected "inside" from the world's "outside", and that membrane does not tear as a result of continuous wheel rotation.

\section{A mechanical introduction to the new system}
\begin{figure}[h]
\begin{center}
\includegraphics[width=8cm]{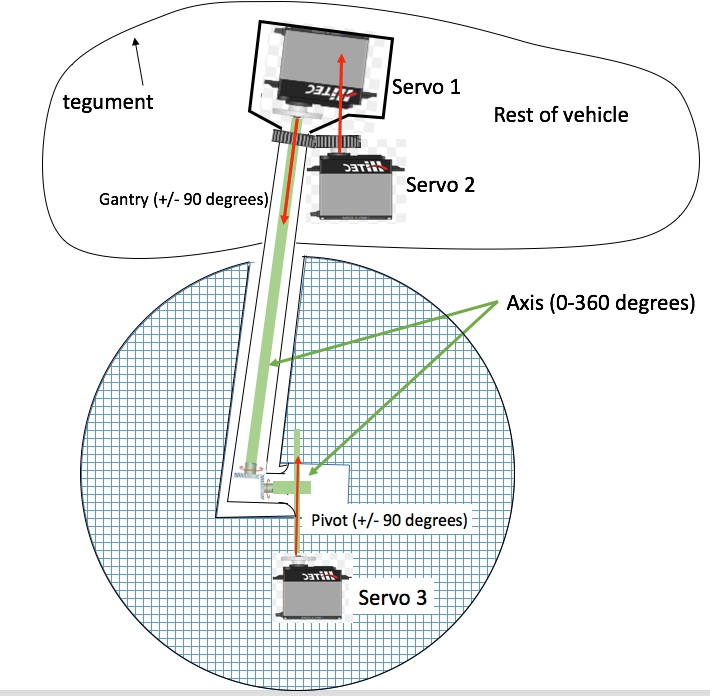}
\caption{Homeostatic wheel, side view. Red arrows indicate positive servo rotations using the right-hand rule.}
\label{hw}
\end{center}
\end{figure}

\begin{figure}[h]
\begin{center}
\includegraphics[width=8cm]{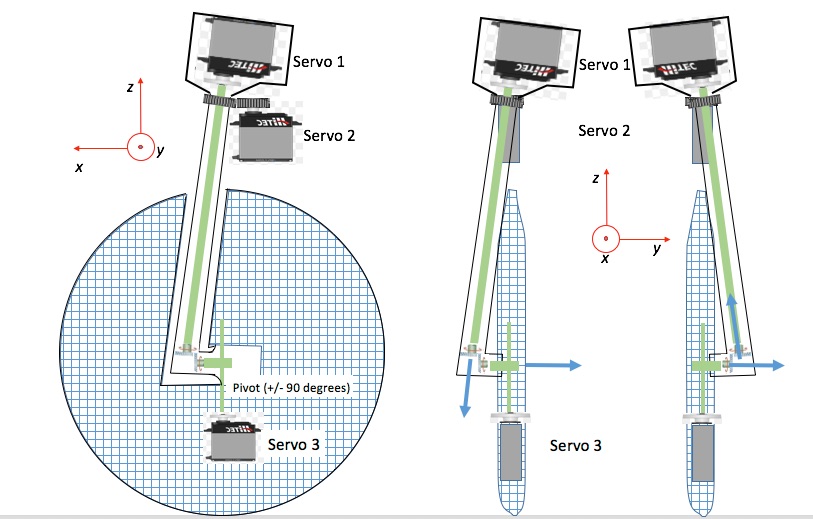}
\caption{Homeostatic wheel, side view and two front views for two opposite positions of the gantry. Each position allow the wheel to rotate under the torque exercised by Servo 1. Right: Blue arrows indicate direction of rotation by employing the right hand rule. Left front view: The gantry is rotated 90 deg to the left, the center shaft rotates as shown, and the wheel rotates forward. Right front view: The gantry is rotated 90 deg to the right, the center shaft rotates as shown -and opposite to the foregoing view- and the wheel still rotates forward.}
\label{hw2}
\end{center}
\end{figure}

The first system description can be constructed through standard electro-mechanical components, such as servos, gears, and linear linkages. The system shown in Fig.~\ref{hw} features the presence of a unique vehicle, whose inside mechanisms are separated from the exterior by a thin deformable membrane which will be referred to as "tegument".
In this implementation, a rigid gantry (shown in black) is wrapped around an articulated transmission shaft. The gantry, which is part of the tegument, and Servo 1 can be rotated relative to the rest of the vehicle via Servo 2, whose rotation angle range is $\left[-90,+90\right]$ degrees. Similarly, servo 3 can rotate the wheel around the green center shaft over the range $\left[-90,+90\right]$, while servo 1 actuates the center shaft (shown in green) and is capable of rotation angles within the range $\left[0,+360\right]$ degrees with no wraparound allowed. i.e. the servo may not rotate more than one full turn total. Front and side views of this apparatus are illustrated in Fig.~\ref{hw2}.
An arbitrary even number of complete forward wheel rotations while keeping tegument integrity is enabled by {\bf Algorithm~\ref{euclid}}. 

{
\begin{algorithm}
\caption{Continuous wheel rotation function.}\label{euclid}\footnotesize
{\bf Function} rotate\_wheel\_2n($n$)\;
{\color{red} \% Assume the wheel is that shown \\\% in Fig.~\ref{hw}, with the angular positions of Servos 1, 2, and 3 angle set to 0\;}
$i := 0$\;
Rotate Servo 3 to -90 degrees\;
Rotate Servo 2 to +90 degrees\;
 {\color{red} \% The configuration is now that shown \\ \% in the middle of Fig.~\ref{hw2}\;}
\While {$i < n$}{
$ i := i+1$ \;
Rotate Servo 1 to 360 degrees\;
{\color{red}  \% The wheel turns a full +360 degree forward \;}
Rotate Servo 3 to +90 degrees\;
Rotate Servo 2 to -90 degrees\;
 {\color{red} \% The configuration is that shown \\ \% to the right of Fig.~\ref{hw2}\;}
Rotate Servo 1 to 0 degrees\;
{\color{red}  \% The wheel turns a full +360 degree forward \;}
Rotate Servo 3 to -90 degrees\;
Rotate Servo 2 to +90 degrees\;
{\color{red} \% The configuration is now that shown \\\% in the middle of Fig.~\ref{hw2}\;}
}
Rotate Servo 3 to 0 degrees\;
Rotate Servo 2 to 0 degrees\;
{\color{red} \% The wheel configuration is that shown \\\% in Fig.~\ref{hw}\;}
{\bf return}\;
\end{algorithm}}

The reader is invited to imagine other combinations of servo rotations to generate arbitrary wheel motions along arbitrary lengths backward and forward, including changes of direction.
Not all axes present in Figs~\ref{hw} and~\ref{hw2} need to be actuated via servos. For example, if the wheel is a free wheel, then servo 1 is unnecessary. Likewise, it might be possible to replace the motions of servos 2 or 3 by passive mechanical links tied to the rolling wheel instead. Additional motorized articulations might be considered to avoid the "gimbal lock" phenomenon that occurs when the angular position of Servos 2 and 3 is zero degrees while the wheel is turning.
We leave it up to the reader to imagine other possible embodiments, some of which may include additional functions, such as suspension systems.

\newpage
\section{An implementation of the system via joints and artificial muscles}
\label{natural_wheel}
A completely consistent and bio-inspired implementation of the continuously rotating wheel presented so far should be based on a combination of artificial muscles and joints. Although it is not exceedingly complex, the description of such a system is necessary to disprove claims that such an appendage could not have resulted from natural evolution of life on Earth: Rather, we claim that the wheel as presented in this paper could have been the result of natural evolution, even in the absence of the accelerations resulting from the development of human intelligence. However, some of the artificial equivalents to natural components may need separate investigations.

\subsection{Skeletal and integumentary elements}
\begin{figure}[h]
\begin{center}
\includegraphics[width=8cm]{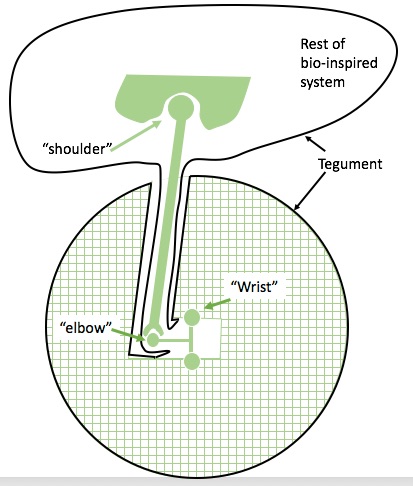}
\caption{Homeostatic wheel: Skeleton (green) and tegument (black).}
\label{hw3}
\end{center}
\end{figure}

The current embodiment is analogous to the foregoing architectural choices and consists of a skeleton, muscles, tendons, and tegument. The skeleton and the tegument are illustrated in Fig.~\ref{hw3}
Although in principle this skeleton representation is feasible, specific issues may arise if the joint angle between any two links ("seen as bones in this section") is too acute, such as in the "elbow" area shown in Fig.~\ref{hw3}. In that case, it is always possible to introduce intermediary links and new joints to enable smoother rotation of all elements involved elements and prevent "gimbal lock". Furthermore, additional skeletal elements may be added to enable the presence of passive equilibria (eg standing upright) as well. In either case, some of these elements may have flexibility built in, allowing for smoother transitions and suspensions.

\subsection{Muscular elements}
The muscles are there to have the skeleton perform all required movements via muscle tension only. It is well-known that muscles/tendons can be arranged, often as antagonistic pairs, to produce various flexing and twisting motions~\cite{PaT:16}. The "shoulder" should be equipped with enough muscles to enable twisting of the center shaft by from zero to 360 degrees as well as overall pivoting of the "leg" by $+/-90$ degrees. The "elbow" should be equipped with enough muscles to enable the transmission of 360 degrees of rotation to the secondary shaft (that between the "elbow" and the "wrist"). Finally, the "wrist" should be equipped with muscles allowing $+/-90$ degrees pivoting of the wheel. The construction of this muscle system heavily depends on available technology, and the acuteness of some of the joint angles may, once again, require the introduction of intermediate skeletal elements to make torque transmission from joint to joint smoother and prevent "gimbal lock". For reviews of existing artificial muscle literature, see~\cite{0964-1726-7-6-001,Bro:91,Ton:15}, for example.

\subsection{Experimental implementation}
The experimental implementation that follows illustrates the ability of the wheel to be part of the "same body" as the vehicle it supports. For logistical reasons, the apparatus does not contain actuation devices (servos or artificial muscles). However, an experimental device that implements such actuation and allows the homeostatic wheel apparatus to act as a motor is currently under implementation according to the foregoing discussion. 
\subsubsection{Skeleton}
\begin{figure}[h]
\begin{center}
\includegraphics[width=5.2cm,angle=90]{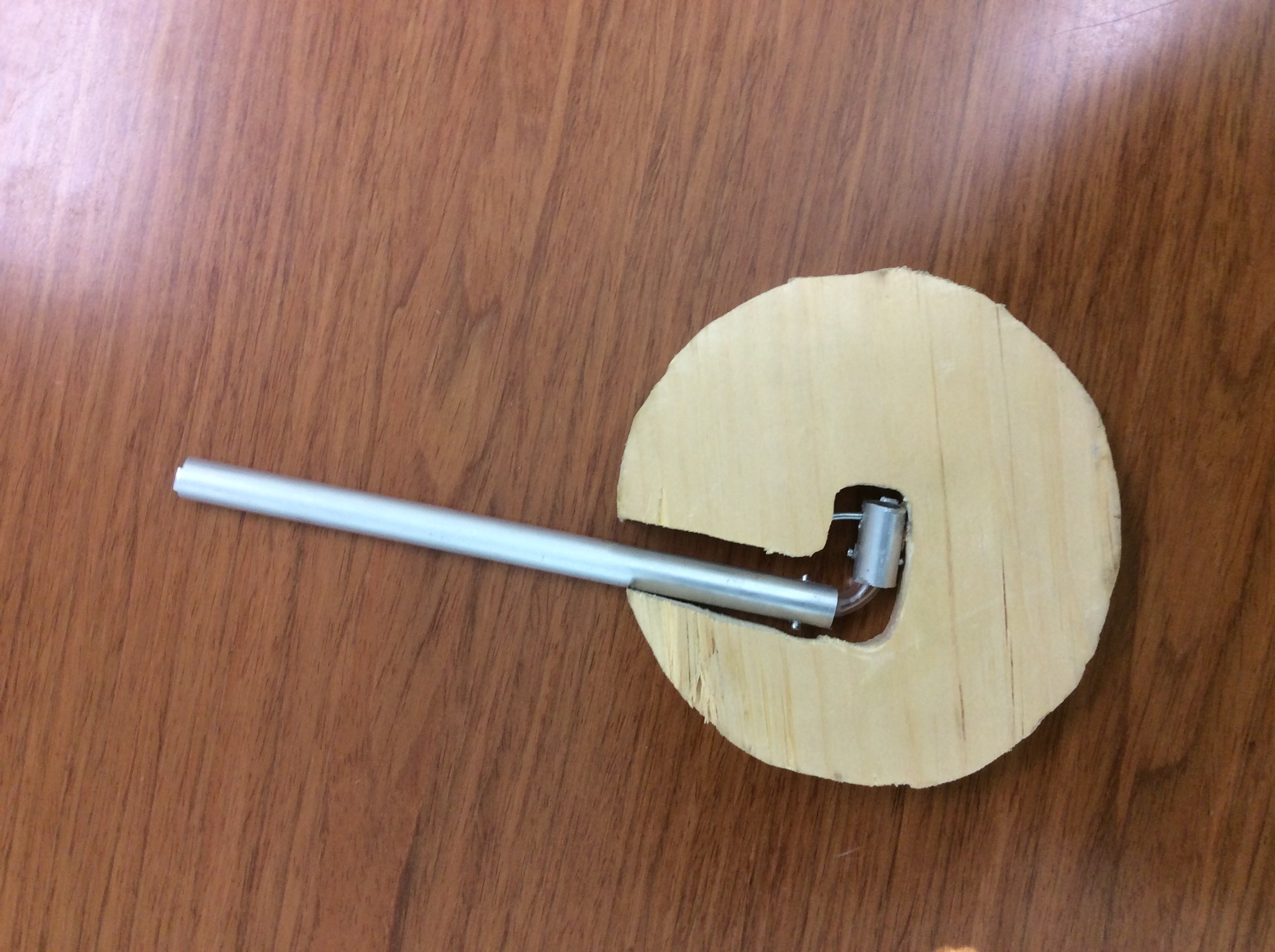}
\includegraphics[width = 4cm]{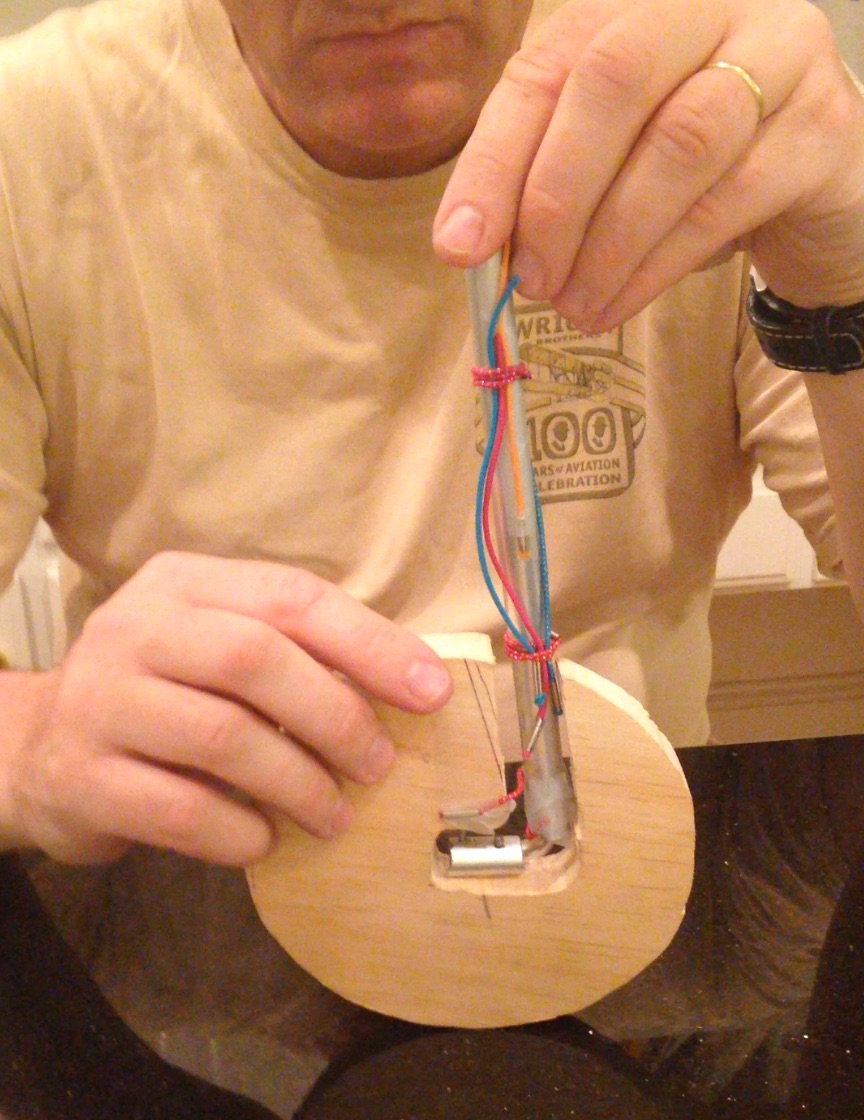}
\caption{Homeostatic wheel implementation: Skeleton. Left: Two pieces of aluminum tube are connected via a flexible link. The second link is connected to the wheel via an axle. Right: Skeleton together with one "blood vessel" connected to the wheel and the supporting rod. The wheel is carved from a piece of wood. A video demonstration of the wheel can be seen at \href{https://youtu.be/3WEJm4lzJRI}{video\_1.mov} }
\label{skeleton}
\end{center}
\end{figure}

First, the skeleton of the system is shown in Fig.~\ref{skeleton}, together with the flexible joint making an "elbow" between two links, and one pivot attaching the second link to the body of the wheel (the wheel is carved from a piece of wood). 
\subsubsection{Skeleton with tegument}
\begin{figure}[h]
\begin{center}
\includegraphics[width=8cm]{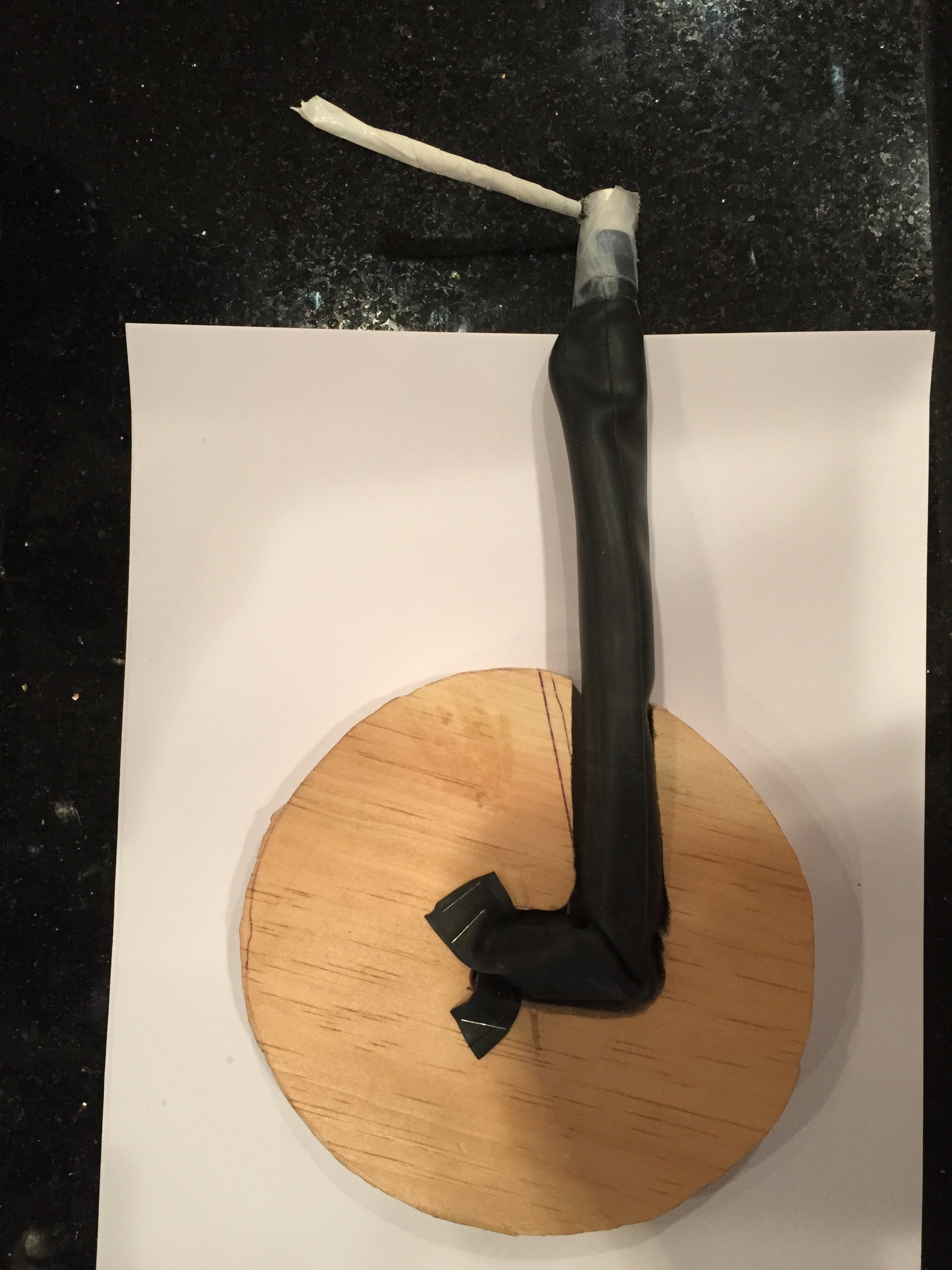}
\caption{Homeostatic wheel implementation: Skeleton and skin. Skin is composed of bicycle tube taped to the shaft, on the one hand, and attached to the wheel outer surface, on the other hand. The wheel "tegument" is the wheel's outer surface. A video demonstration of the wheel with tegument can be seen at \href{https://youtu.be/5k5vOYvvIyI}{video\_2.mov}}
\label{skin}
\end{center}
\end{figure}

Next, the system was fitted with a "skin" made of a bicycle tire tube, as shown in Fig.~\ref{skin}.
The tube, together with the surface of the wheel, forms a continuous tegument that wraps both the transmission shafts and the wheel. 

\section{Discussion}

\subsection{Mathematical considerations}
\subsubsection{Relation to Dirac's belt trick}
The concept described in this paper is a direct application of "Dirac's Belt trick"~\cite{Sta:31}, \cite[p. 346]{Lom:00} (see also {\tt https://en.wikipedia.org/wiki/Plate\_trick}), and is closely related to the group of rotations $SO(3)$, the group $SU(2)$, and quaternions. From that perspective, the wheel and the surrounding system described in this paper is a "rigid" realization of Dirac's belt, and the angles used for all servos involved are in many ways similar to Euler angles, and bring with them traditional gimbal lock problems when Servos 2 and 3 are set at zero and the wheel is rotating. Adams~\cite{Ada:98} also evokes Dirac's belt trick by defining an "Umbilical cord" when discussing nature's inability to come up with a wheel, though he does not proceed with a full description of how the system would work. Dirac's belt trick has been used in many applications, though not for that described here. One of them is Adams' most noteworthy invention~\cite{adams1971apparatus}, also named "anti-twister mechanism".

\subsubsection{Relation to control theory}
From a control systems perspective, the system described here illustrates the strong possibilities raised by motion rectification, whereby oscillations of key elements in an articulated system transform into continuous motion of other elements (in our case, the wheel). There is a considerable corpus of mathematical knowledge for such systems, that belong to the class of {\em non holonomic} systems. The mathematical foundations of how to operate these systems can be traced at least as far back as Brockett and co-workers~\cite{Bro:81,Bro:89b,Bro:87,Bro:89,kliman1991vibratory}, and have been the subject of many investigations ever since (see~\cite{Blo:15} for a comprehensive overview). In particular, there are likely implementations of the non-holonomic system discussed here that rely on the smooth integration of appropriately phased oscillations of the various muscles / servos described here to result in smooth continuous rotation of the wheel or other rotating elements. The same considerations have also been pointed out by R. Brooks and co-authors~\cite{Bro:02,Bro:89}
for example. It is then possible to imagine that full leveraging of the mechanism described here could be performed via online behavior learning techniques~\cite{MaB:90}.

\subsection{Wheel integrity}
\begin{figure}[h]
\begin{center}
\includegraphics[width=8cm]{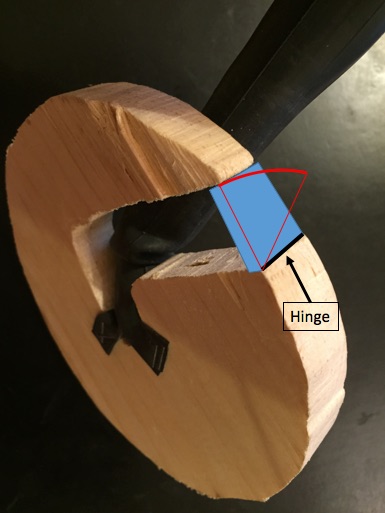}
\caption{Schematic of mechanism to recover full wheel shape. This mechanism can be implemented in an homeostasis-enabling way.}
\label{cont_wheel}
\end{center}
\end{figure}

From the description above, one might argue that the proposed system is not exactly a "wheel", since it must be cut to allow the torque transmission mechanism to switch freely from one side of the wheel to the other. This imperfection can be solved using servos or artificial muscles by adding a movable part that closes the wheel upon contact with the ground, and opens up when the shaft must rotate from one side of the wheel to the other as shown, for example, in Fig.~\ref{cont_wheel}. However, this fix may not solve more important issues, such as the difficulty to wrap a classical tire around this wheel. For that purpose specific research efforts need performing.
As for the rest of the system, there exist simple implementations of the wheel integrity mechanism that preserve homeostasis of the entire vehicle (supported platform + wheels): Simply wrap a tegument around the hinge in Fig.~\ref{cont_wheel}, or make the entire hinged mechanism part of the  flexible tegument.

\subsection{Applications}
The range of potential applications of the proposed mechanism includes all current applications of actuated rotating systems, be they wheel, propellers, or other. One particularly interesting aspects of the work arises towards small sizes, especially when artificial muscles are considered for implementation. Indeed, for any given wheel of size $L$ (measured in units of length), the mass of the wheel spinning system grows approximately like $L^3$ and the strength of the muscles involved grows like their cross section, that is a quantity proportional to $L^2$. The ratio of applicable forces to system mass is crucial in system dynamics. For example, Newton's law
\[
F = m a, \mbox{ or } a = \frac{F}{m}
\]
shows that possible accelerations depend directly on this ratio, which is proportional to $1/L$. When $L$ is large, possible accelerations decrease in magnitude, but when $L$ is small (small system dimensions), the possible accelerations of all elements of the system increase significantly. Eventually, the level of success of the implementation of the proposed wheel system could be evaluated using Diamond's "cost of transport" metric~\cite{Dia:83}, for example.

\section{Evolutionary considerations}
The wheel presented in this document, especially that in Section~\ref{natural_wheel} completely respects the constraints limiting evolution of animal limbs to respecting the basic topological constraints for 
homeostasis to exist. Intuition therefore suggests that a continuous evolution could have occurred in animals at diverse degrees of evolution to eventually develop and use such "wheels". However, drawing a plausible development path would require a collaboration with specialists in the field. In particular, one requirement for this path is that all intermediate evolutionary states from initial state to final, "wheeled" state be "usable" and "desirable".

\section*{Conclusion}
A new kind of mechanism has been introduced to replicate the behavior of wheels while preserving the kind of "system interior" that, alone, can enable homeostasis and the enabling of life-like self-repair and energy management mechanisms. Various implementations of the mechanism have been discussed, and a simple prototype has been built that illustrates the point of this paper. The full range of applications of the proposed system remains to be evaluated.

\bibliography{biblio}

\end{document}